%% file: neurips_2025.tex
\documentclass{article}

\usepackage{neurips_2025}

\usepackage[utf8]{inputenc} % allow utf-8 input
\usepackage[T1]{fontenc}    % use 8-bit T1 fonts
\usepackage{hyperref}       % hyperlinks
\usepackage{url}            % simple URL typesetting
\usepackage{booktabs}       % professional-quality tables
\usepackage{amsfonts}       % blackboard math symbols
\usepackage{nicefrac}       % compact symbols for 1/2, etc.
\usepackage{microtype}      % microtypography

\newcommand{\et}[2]{${#1}^{\pm{#2}}$}

\usepackage[inkscapelatex=false]{svg}

\usepackage{microtype}
\usepackage{url}
\usepackage{times}
\usepackage{latexsym}
\usepackage{booktabs}
\usepackage{multirow}
\usepackage{multicol}
\usepackage[mathscr]{euscript}
\usepackage{xcolor}
\usepackage{colortbl}
\usepackage{makecell}
\usepackage{subcaption}
\usepackage{inconsolata}

\usepackage{graphicx}

\input{math_commands.tex}

\usepackage{times}
\usepackage{amsmath} 

\usepackage{booktabs}
\usepackage{array}
\usepackage{caption}

\usepackage{algorithm}
\usepackage{multirow}
\usepackage{bm}
\usepackage{graphicx}
\usepackage{color}
\usepackage{booktabs}
\usepackage{arydshln}
\usepackage{amsfonts}
\usepackage{eqnarray}
\usepackage{stfloats}
\usepackage{wrapfig}

\usepackage{amsmath} 
\usepackage[utf8]{inputenc} % allow utf-8 input
\usepackage[T1]{fontenc}    % use 8-bit T1 fonts
\usepackage{hyperref}       % hyperlinks
\usepackage{url}            % simple URL typesetting
\usepackage{booktabs}       % professional-quality tables
\usepackage{amsfonts}       % blackboard math symbols
\usepackage{nicefrac}       % compact symbols for 1/2, etc.
\usepackage{microtype}      % microtypography
\usepackage{graphicx}
\usepackage{multirow}

\newcommand{\modelname}{PlanMoGPT}

\title{\modelname: Flow-Enhanced Progressive Planning \\ for Text to Motion Synthesis}

\newcommand*{\affaddr}[1]{#1} % No op here. Customize it for different styles.
\newcommand*{\affmark}[1][*]{\textsuperscript{#1}}

\author{%
Chuhao Jin\affmark[1], Haosen Li\affmark[1], Bingzi Zhang\affmark[1], Che Liu\affmark[2], Xiting Wang\affmark[1], Ruihua Song\affmark[1]\thanks{Corresponding authors: Ruihua Song (rsong@ruc.edu.cn).},\\
\textbf{Wenbing Huang\affmark[1], Ying Qin\affmark[1], Fuzheng Zhang\affmark[2], Di Zhang\affmark[2]}\\
\normalsize
\affaddr{\affmark[1]Renmin University of China, Beijing, China}\\
\affaddr{\affmark[2]Kuaishou, Beijing, China}
}

\begin{document}

\maketitle

\input{sec/abs}
\input{sec/intro}
\input{sec/related_work}
\input{sec/method}
\input{sec/experiment}
\input{sec/conclusion}

\bibliographystyle{plain}
\bibliography{main}

%%%%%%%%%%%%%%%%%%%%%%%%%%%%%%%%%%%%%%%%%%%%%%%%%%%%%%%%%%%%

\appendix
\input{sec/appendix}
% \input{sec/limitation_ethic}
% \input{sec/checklist}

%%%%%%%%%%%%%%%%%%%%%%%%%%%%%%%%%%%%%%%%%%%%%%%%%%%%%%%%%%%%

\end{document}

%% file: math_commands.tex
\usepackage{amsmath,amsfonts,bm}

\def\eqref#1{equation~\ref{#1}}
\def\1{\bm{1}}

\DeclareMathAlphabet{\mathsfit}{\encodingdefault}{\sfdefault}{m}{sl}
\SetMathAlphabet{\mathsfit}{bold}{\encodingdefault}{\sfdefault}{bx}{n}

%% file: sec/abs.tex
\begin{abstract}
Recent advances in large language models (LLMs) have enabled breakthroughs in many multimodal generation tasks, but a significant performance gap still exists in text-to-motion generation, where LLM-based methods lag far behind non-LLM methods.
We identify the granularity of motion tokenization as a critical bottleneck: fine-grained tokenization induces local dependency issues, where LLMs overemphasize short-term coherence at the expense of global semantic alignment, while coarse-grained tokenization sacrifices motion details.
To resolve this issue, we propose \modelname, an LLM-based framework integrating progressive planning and flow-enhanced fine-grained motion tokenization. First, our progressive planning mechanism leverages LLMs' autoregressive capabilities to hierarchically generate motion tokens by starting from sparse global plans and iteratively refining them into full sequences. Second, our flow-enhanced tokenizer doubles the downsampling resolution and expands the codebook size by eight times, minimizing detail loss during discretization, while a flow-enhanced decoder recovers motion nuances. Extensive experiments on text-to-motion benchmarks demonstrate that \modelname~achieves state-of-the-art performance, improving FID scores by 63.8\% (from 0.380 to 0.141) on long-sequence generation while enhancing motion diversity by 49.9\% compared to existing methods. The proposed framework successfully resolves the diversity-quality trade-off that plagues current non-LLM approaches, establishing new standards for text-to-motion generation. Project page: \url{https://PlanMoGPT.github.io}.
\end{abstract}

%% file: sec/intro.tex
\section{Introduction}
\label{sec:intro}

Large language models (LLMs) have demonstrated remarkable capabilities across diverse tasks~\cite{brown2020language}, including multimodal scenarios~\cite{liang2024survey,achiam2023gpt,alayrac2022flamingo,liu2023visual}, by leveraging world knowledge and reasoning abilities acquired through large-scale pretraining~\cite{wei2022emergent,kojima2022large}. Their capacity for autoregressive planning and commonsense reasoning has proven critical for complex multimodal generation tasks~\cite{zhao2023large,wang2024exploring}, such as image generation~\cite{koh2023generating} and speech synthesis~\cite{zhang2023speechgpt}. However, in text-to-motion generation, LLM-based methods still underperform compared to non-LLM approaches (e.g., diffusion-based methods~\cite{tevet2023human,zhang2022motiondiffuse,zhang2023remodiffuse,chen2023executing,shafir2024human}), although the latter still faces limitations such as slow inference speed and low diversity. This paradox motivates our investigation into a fundamental research question: What limits the potential of LLMs in motion generation?
Addressing this bottleneck may overcome the limitations of current text-to-motion generation approaches while establishing a new paradigm for efficient, high-quality motion generation.

\begin{figure}
    \centering
    \includegraphics[width=1.0\linewidth]{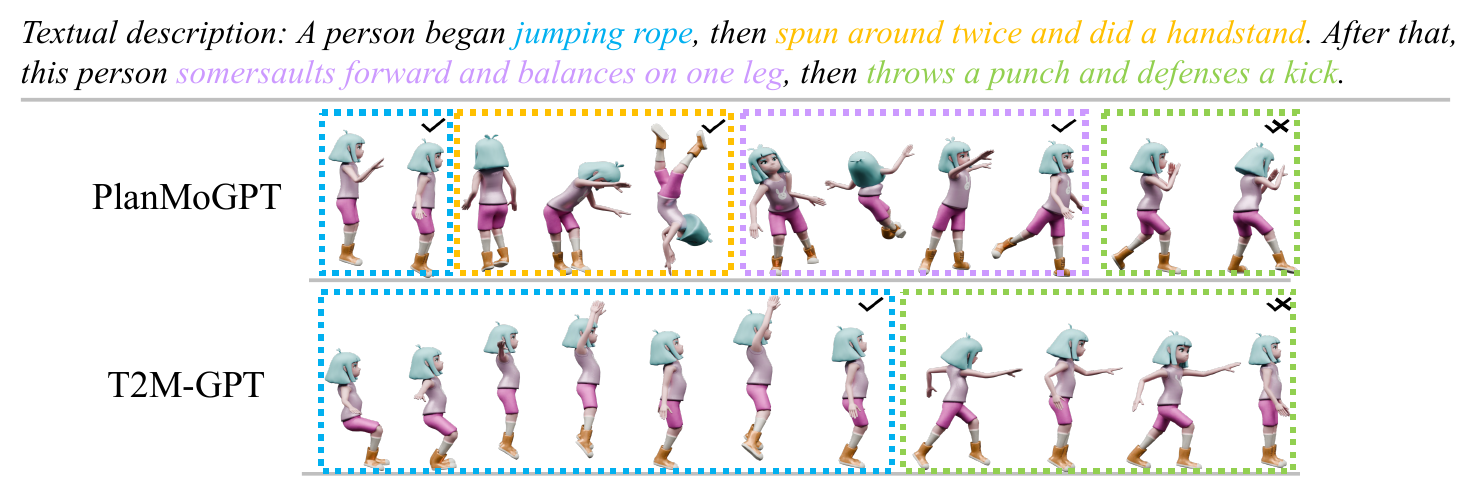}
    \caption{Generating complex and long-sequence motion by our~\modelname, T2M-GPT~\cite{zhang2023generating}. The bounding box of each motion clip is color-coded to match its corresponding text.
    }\label{fig:teaser}\vspace{-18pt}
\end{figure}

We identify the granularity of motion tokenization as a core challenge. 
Specifically, previous LLM-based motion generation approaches typically follow the token-based paradigm. 
In this paradigm, it first uses a motion tokenizer (e.g., VQ-VAE~\cite{van2017neural}) to compress motion sequences into discrete tokens.
LLMs are then trained to generate these tokens, and finally, a decoder is used to reconstruct the motion from generated tokens. 
Although this approach can incorporate continuous motion into the LLM generation paradigm, motion is represented as a sequence of fine-grained tokens, and similar motions are divided into similar tokens. This causes adjacent tokens to be very similar, which makes the newly generated token provide a strong contextual signal for the next token prediction, thus letting the model ignore the text and earlier tokens, leading to a local dependency problem.
As evidenced in Figure~\ref{fig:teaser}, for a complex and long motion description, T2M-GPT~\cite{guo2022generating} frequently exhibits partial action generation (too much jumping but omitting other actions), as it overemphasizes local motion coherence at the expense of global alignment. 
Although coarse-grained tokenization can alleviate the local dependency problem, this will lead to the loss of motion details~\cite{zhang2023generating} and even make the generated motion discontinuous.
This shows that there is a conflict between preserving motion details in motion discretization and generating token sequences stably by LLMs.

To address the above challenges, we propose an LLM-based motion generation framework named \modelname, which integrates an LLM-based progressive planning mechanism with flow-enhanced fine-grained motion tokenization.
First, we resolve local dependency limitations in fine-grained generation through the proposed progressive planning mechanism. Leveraging LLMs' planning capabilities, the system first generates motion tokens at large intervals (e.g., every 4 frames) to establish global motion structure, as a preliminary plan. This plan is then progressively refined to middle-grained (2-frame intervals) and until full-sequence through iterative prediction. 
Our experimental results show that this progressive planning enables the LLM to generate high-fidelity fine-grained token sequences, even when handling long-sequence motion generation tasks.
Building on this foundation, we develop a flow-enhanced fine-grained motion tokenizer with minimal detail loss.
Firstly, we employ the VQ-VAE to encode the motion, increasing the codebook size to eight times while reducing the downsampling rate by half compared to prior works~\cite{zhang2023generating,jiang2023motiongpt}. 
During motion decoding, we employ a flow-enhanced motion decoder to establish gradient-based mapping from coarse to refined motions, effectively recovering motion details lost during tokenization. 
Experimental results indicate that our proposed~\modelname~significantly outperforms all LLM-based and non-LLM baselines in all metrics, including generation quality, semantic alignment, and diversity, on multiple text-to-motion datasets. This multi-interval LLM-based planning approach also balances high diversity with good semantic alignment simultaneously.

Our contributions are threefold: 
\begin{itemize}
    \item We propose a text-to-motion generation framework named~\modelname, which better leverages LLMs by addressing the challenge of the granularity of motion tokenization;
    \item We propose a progressive planning method to address the local dependency problem in LLMs, and a flow-enhanced fine-grained motion tokenizer to encode finer-grained motions;
    \item Experimental results indicate that~\modelname~achieves state-of-the-art performance across multiple text-to-motion benchmarks. Particularly significant improvements are observed on long-sequence datasets, where FID improves from 0.380 to 0.141. Experimental results show that \modelname~resolves the diversity-quality dilemma in existing non-LLM approaches. It generates highly diverse motions, with a 49.9\% improvement in diversity metrics over the state-of-the-art baselines, while maintaining superior generation quality in FID scores.
\end{itemize}

%% file: sec/related_work.tex
\section{Related Work}
\label{sec:related}

\subsection{Human Motion Synthesis}
Human motion synthesis aims to generate semantic-aligned and realistic motions conditioned on multimodal inputs, including texts~\cite{zhu2023human,guo2022generating,hong2022avatarclip,guo2024momask,zhang2023finemogen,zhang2022motiondiffuse,zhang2023remodiffuse,pinyoanuntapong2024bamm}, action labels~\cite{yu2020structure,degardin2022generative,guo2020action2motion,petrovich2021action}, scene contexts~\cite{cao2020long,wang2021scene,hassan2021stochastic,taheri2022goal,wu2022saga}, speech signals~\cite{ao2023gesturediffuclip,yang2023qpgesture,kucherenko2019analyzing,li2021audio2gestures,liu2022beat}, and partial motion sequences~\cite{li2021ai,punnakkal2021babel,ginosar2019learning,chang2017matterport3d}. Recent text-to-motion approaches primarily employ four generative paradigms: GANs~\cite{goodfellow2014generative,xu2023actformer,men2022gan}, VAEs~\cite{kingma2013auto,van2017neural}, diffusion models~\cite{ho2020denoising,weng2021diffusion,song2020score}, and token-based~\cite{zhang2023generating,guo2024momask,pinyoanuntapong2024bamm,wu2025motionagent}, with diffusion and token-based methods emerging as dominant trends. To align generated motions with texts well, ReMoDiffuse~\cite{zhang2023remodiffuse} guides the generation through retrieved reference motion, but this may introduce retrieval model bias. FineMoGen~\cite{zhang2023finemogen} introduces time-aligned text decomposition but may compromise global motion coherence. 
However, the iterative generation in diffusion methods leads to inefficient inference. Recent acceleration attempts use Flow Matching~\cite{hu2023motion,lipman2022flow} for faster inference but sacrifice motion quality. 
Compared with previous works, we address the granularity of motion tokens. We propose the LLM-based progressive planning with a fine-grained flow-enhanced motion tokenizer, which not only delivers state-of-the-art generation quality but also demonstrates substantially faster inference speeds compared to mainstream diffusion models.

\subsection{Text and Motion Alignment}
The goal of text and motion alignment is to achieve an accurate understanding of discrete language and continuous motion,
so as to benefit various downstream tasks, such as text-to-motion~\cite{pinyoanuntapong2024bamm,ao2023gesturediffuclip,alayrac2022flamingo}, motion caption~\cite{wang2024motiongpt,wu2025motionagent}.
Recent studies use the CLIP~\cite{radford2021learning} model to align texts and motions, e.g., MotionCLIP~\cite{tevet2022motionclip} aligns text and motion images to the latent space of CLIP~\cite{radford2021learning}, T2M-GPT~\cite{zhang2023generating} and MotionDiffuse~\cite{zhang2022motiondiffuse} use CLIP's text encoder to enhance visual understanding, etc. But it is difficult for the image-based CLIP model to achieve precise alignment at the temporal level.
Token-based methods~\cite{wang2024motiongpt,pinyoanuntapong2024bamm,zhang2023generating,wu2025motionagent} project motion into the language space, thus showing a superior alignment performance. MotionGPT~\cite{jiang2023motiongpt} uses post-training tasks to further promote convergence.
Yet they suffer from detail loss in quantization processes. Although MoMask~\cite{guo2024momask} addresses quantization losses via multi-layer residual learning, its hierarchical structure distributes motion details across various layers, ultimately weakening base model capabilities and yielding suboptimal results. 
Compared with previous work, our fine-grained motion tokenizer retains more details in tokens, and the progressive planning aligns the motion and text at different granularities.

%% file: sec/method.tex
\section{Method}  
\label{sec:method}  
Given a textual description $x$, our goal is to generate a high-quality and semantically aligned motion $y = \{p_i\}_{i=1}^{n}$,
where each pose $p_i \in \mathbb{R}^{d_m}$ is a $d_m$-dimensional vector capturing the position information of human joints at the i-th frame. 
As shown in Figure~\ref{fig:main_method}, our \modelname~consists of two parts: Flow-enhanced motion tokenizer converts motion into token sequence with minimal detail loss (Sec \ref{sec:vqvae}), and an LLM integrate with progressive planning for token sequence generation (Sec \ref{sec:plan}).

\subsection{Flow-Enhanced Motion Tokenizer}
\label{sec:vqvae}
As the prerequisite stage of our framework, the tokenizer must first establish an accurate discrete motion representation for the following LLM-based motion generation. 
We first introduce the baseline approach and its limitations, and then introduce the motion tokenizer we use, including fine-grained motion encoding and flow-enhanced motion decoding.

\noindent\textbf{Baseline Architecture.} Previous approaches~\cite{jiang2023motiongpt} process input motion $y \in \mathbb{R}^{n \times d_m}$ through a CNN-based encoder with stride $r=4$, producing latent vectors $Z=\{z_i\}_{i=1}^{l}$, where $l=\lfloor n/r \rfloor$. These are quantized via nearest-neighbor lookup in a codebook $\mathcal{C} \in \mathbb{R}^{K \times d}$ (typically $K=512$), generating discrete token sequence $M=\{m_i\}_{i=1}^l$. 
Then, through a CNN-based decoder, $M$ is reconstructed into motion sequence, which can be denoted as $y_0 \in \mathbb{R}^{n \times d_m}$.
While effective for representation, this paradigm suffers from two limitations: 1) High downsampling rates discard temporal resolution, and 2) Small codebooks restrict expressiveness, causing irreversible loss of motion details during quantization.
Although previous works have also attempted to use more fine-grained tokens, such as T2M-GPT~\cite{zhang2023generating} using a larger codebook, this even causes generated motion quality deterioration, due to downstream generation models lacking sufficient capacity.

\begin{figure}
    \centering
    \includegraphics[width=1.0\linewidth]{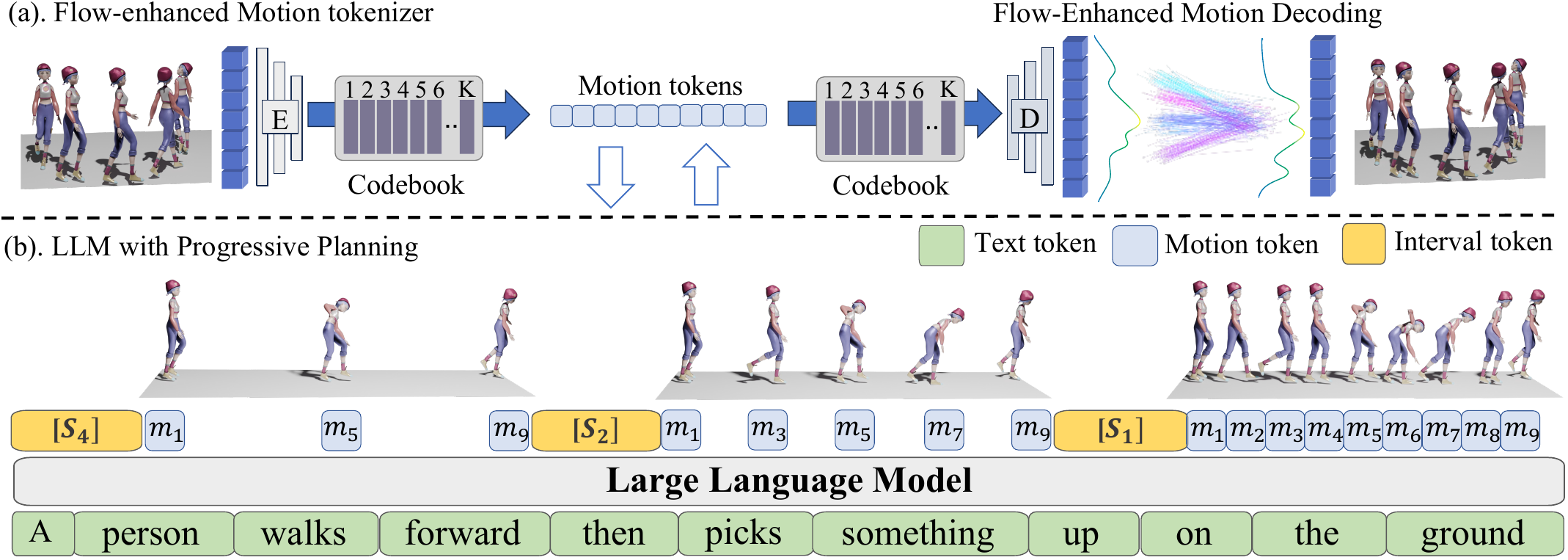}
    \caption{\modelname~consists of two components: (a). A flow-enhanced motion tokenizer converts motion into fine-grained tokens with minimal loss; (b). An LLM integrates with progressive planning, which progressively generates from a larger interval motion tokens to the full motion token sequence.}\vspace{-10pt}
    \label{fig:main_method}
\end{figure}

\noindent\textbf{Fine-Grained Motion Encoding.} 
We improve the previous VQ-VAE through two improvements. First, we reduce the downsampling rate to $r=2$.
This preserves a two-times temporal resolution compared to conventional approaches.
To better capture motion details, we incorporate additional convolutional layers with a stride of 1, maintaining fidelity throughout the encoding process.
Secondly, we enlarge the codebook size to 4096, eight times larger than previous works.
The enlarged code space can represent more motion details, thereby reducing the quantization loss.

\noindent\textbf{Flow-Enhanced Motion Decoding.}
While the improved motion encoding enables fine-grained motion tokenization, its CNN-based decoder's output $y_0$ still exhibits details losses due to quantization.
To bridge this quality gap, we propose to refine $y_0$ through flow matching~\cite{lipman2022flow} – a continuous optimization process that progressively adds details to the coarse ``motion sketch.''
Specifically, as shown in~Figure \ref{fig:main_method}, we leverage the flow-matching method to establish a smooth optimization trajectory that evolves the $y_0$ into the real motion sequence $y$ (now denoted as $y_1$). Guided by a learnable vector field \( \mathbf{F}_\theta \), this process is formalized as an ordinary differential equation (ODE): 
\begin{equation}
    \frac{dy_t}{dt} = \mathbf{F}_\theta(y_t, t), \quad t \in [0,1]
\end{equation}
Here, $t$ serves as a virtual time variable: $y_t = y_0$ (coarse motion) at $t=0 $, and $y_t = y_1$ (real motion) at $t=1$. The vector field $\mathbf{F}_\theta$ acts as a correction signal generator, predicting both the direction and magnitude of adjustments needed for the current motion sequence $y_t$.
We optimize $\mathbf{F}_\theta$ using conditional flow matching:
\begin{equation}
    \mathcal{L}_{\text{CFM}}(\theta) = \mathbb{E}_{t, y_1, y_t} \big\| \mathbf{F}_\theta(y_t, t) - u(y_t|y_1) \big\|^2,
\end{equation}
where $u(y_t|y_1)$ represents the tangent direction of the shortest path between $y_t$ and $y_1$. This loss function forces the network to learn the most efficient incremental corrections.

In inference, we discretize the ODE into $T$ refinement steps. Starting from $y_0$, we iteratively update:
\begin{equation}
    y_{t_{i+1}} = y_{t_i} + \mathbf{F}_\theta(y_{t_i}, t_i)\cdot\Delta t, 0 = t_0 < \cdots < t_T = 1
\end{equation}
where $\Delta t$ is the time increment. This iterative scheme progressively injects motion details (e.g., natural limb oscillations) into the motion sequence $y_0$.

\subsection{LLM with Progressive Planning}
\label{sec:plan}
Building on our flow-enhanced motion tokenizer, we design an LLM-based progressive planning mechanism to address the local dependency problem in fine-grained token generation. Our approach consists of two components: Multi-interval plan sampling and a progressive generation pipeline.

\noindent\textbf{Multi-Interval Plan Sampling.} 
Given a motion token sequence $M = \{m_i\}_{i=1}^{l}$, we first construct multi-granularity motion plans through interval sampling:
\begin{equation}
M_{b;T} = \{m_{b+kT} | 0 \leq k \leq \lfloor (l-b)/T \rfloor\},
\end{equation}
where $T \in \{4,2,1\}$ controls temporal granularity (larger $T$ indicates coarser plans). The base offset $b$ is randomly sampled from $\{1,..,T_{\text{max}}\}$ to ensure diverse plan initialization. This creates a three-level hierarchy: 4-frame interval plans for global structure, 2-frame plans for intermediate refinement, and full-sequence tokens for detail synthesis.

\noindent\textbf{Progressive Generation Pipeline.} 
We reformulate motion generation as a hierarchical sequence prediction task by combining multi-interval plans as follows:
\begin{equation}
U = [ \text{S}_4 ] \oplus M_{b;4} \oplus [ \text{S}_2 ] \oplus M_{b;2} \oplus [ \text{S}_1 ] \oplus M,
\end{equation}
where $\oplus$ denotes concatenation and $[ \text{S}_i ]$ are special tokens marking granularity transitions.

This structure enables the following advantages: 

\noindent1) \textbf{Coarse-to-Fine Generation}: The 4-frame plans ($T=4$) force the LLM to focus on global semantics by removing 75\% of surrounding tokens, establishing text-aligned motion skeletons. Then the 2-frame plans ($T=2$) inject mid-scale kinematics while maintaining semantic consistency through cross-attention with coarser plans, and finally, the full-sequence generation ($T=1$) focuses on detail completion under dual constraints from both upper levels; 

\noindent2) \textbf{Cross-Level Error Correction}: 
The generation of full sequence benefits from both upper-level plans, with cross-attention weights showing 32.9\% and 15.3\% allocation to $M_{b;4}$ and $M_{b;2}$.
By maintaining continued access to the plans, cumulative errors in fine-grained generation can be rectified through backward consistency checks with coarser plans. 

Overall, this progressive planning mechanism transforms motion generation from conventional token-by-token prediction to structured plan evolution. As shown in our experiments, the hierarchy-aware generation process achieves better long-term consistency than standard autoregressive approaches while maintaining the detail-preserving capability of fine-grained tokenization.

\begin{table*}[t]
\centering
\label{tab:dataset_stats}
\scalebox{0.8}{\begin{tabular}{lrrrrrrrr}
\toprule
Dataset & Motions & Texts & \makecell{Texts per \\ Motion} & \makecell{Frame Rate \\ (FPS)} & \makecell{Avg. Duration \\ (s)} & \makecell{Max Duration \\ (s)} & Vocab Size & \\
\midrule
KIT-ML & {3,911} & {6,278} & 1--4 & 12.5 & 6.1 & 10 & {1,623}\\
KIT-ML++ & 24,234 & 25,833 & 1--4 & 12.5 & 27.3 & 50 & 2,297 \\\midrule
HumanML3D & {14,616} & {44,970} & 3 & 20 & 7.1 & 10 & {5,317} \\
HumanML3D++  & {127,990} & 143,728 & 1--3 & 20 & 20.1 & 50 & 8,593 \\
\bottomrule
\end{tabular}
}\caption{Statistics of the text-to-motion generation datasets.}\vspace{-10pt}
\label{tab:data_stat}
\end{table*}

%% file: sec/experiment.tex
\section{Experiments}
\label{sec:exper}

\begin{table*}[t]
    \centering
    \scalebox{0.8}{
    \begin{tabular}{l l c c c c c c}
    \toprule
\multirow{2}{*}{Datasets} & \multirow{2}{*}{Methods}  & \multicolumn{3}{c}{R-Precision $\uparrow$} & \multirow{2}{*}{FID $\downarrow$} & \multirow{2}{*}{MM-Dist $\downarrow$} & \multirow{2}{*}{MModality $\uparrow$}\\
    \cline{3-5}
    ~ & ~ & Top-1 & Top-2 & Top-3 \\\midrule
\multirow{13}{*}{\makecell[l]{Human\\ML3D}} &  MDM$^\S$~\cite{mahmood2019amass} & - & - & \et{0.611}{.007} & \et{0.544}{.044} & \et{5.566}{.027} & \et{2.799}{.072}  \\ 
~&    MFM$^\S$~\cite{hu2023motion} & - & - & \et{0.642}{.003} & \et{0.362}{.006} & \et{5.280}{.009}  & \et{2.443}{.070}  \\ 
~&    MotionDiffuse$^\S$~\cite{zhang2022motiondiffuse} & \et{0.491}{.001} & \et{0.681}{.001} & \et{0.782}{.001} & \et{0.630}{.001} & \et{3.113}{.001}  & \et{1.553}{.042}  \\ 
~&    ReMoDiffuse$^\S$~\cite{zhang2023remodiffuse} & \et{0.510}{.005} & \et{0.698}{.006} & \et{0.795}{.004} & \et{0.103}{.004} & \et{2.974}{.016} &\et{1.795}{.043}\\
\cline{2-8}
~&    T2M-GPT~\cite{zhang2023generating} & \et{0.491}{.003} & \et{0.680}{.003} & \et{0.775}{.002} & \et{0.116}{.004} & \et{3.118}{.011} &  \et{1.856}{.011} \\
~&    T2M-GPT* & \et{0.494}{.003} & \et{0.684}{.002} & \et{0.777}{.003} & \et{0.130}{.004} & \et{3.109}{.006} &   \et{2.363}{.651} \\
~ & MotionGPT~\cite{jiang2023motiongpt} & \et{0.492}{.003} & \et{0.681}{.003} & \et{0.778}{.002} & \et{0.232}{.008} & \et{3.096}{.008} &\et{2.008}{.084}\\
~ & MotionLLM~\cite{wu2025motionagent} & \et{0.515}{.004} & - & \et{0.801}{.004} & \et{0.230}{.009} & \et{2.967}{.020} & - \\
~&    MoMask (base) $^\S$ & \et{0.504}{.004} & \et{0.699}{.006} & \et{0.797}{.004} & \et{0.082}{.008} & \et{3.050}{.013} &\et{1.050}{.061}\\
~&    MoMask$^\S$~\cite{guo2024momask} & \et{0.521}{.002} & \et{0.713}{.002} & \et{0.807}{.002} & \et{\textbf{0.045}}{.002} & \et{{2.958}}{.008} & \et{1.241}{.040}\\
~&    BAMM (base) $^\S$ & \et{0.507}{.003} & \et{0.703}{.002} & \et{0.799}{.002} & \et{0.111}{.005} & \et{3.028}{.009}  &\et{1.750}{.057}\\
~&    BAMM$^\S$~\cite{pinyoanuntapong2024bamm} & \et{\underline{0.522}}{.002} & \et{\underline{0.715}}{.002} & \et{\underline{0.808}}{.002} & \et{{0.055}}{.002} & \et{\underline{2.936}}{.008}  &\et{1.732}{.055}\\
\cline{2-8}
~ & \modelname~(base) & \et{0.521}{.003} & \et{0.711}{.002} & \et{0.804}{.002} & \et{0.106}{.004} & \et{{2.937}}{.008}  &\et{\underline{2.814}}{.235}\\
~ &  \textbf{\modelname~} & \et{\textbf{0.526}}{.002} & \et{\textbf{0.716}}{.002} & \et{\textbf{0.809}}{.002} & \et{\underline{0.048}}{.002} & \et{\textbf{2.884}}{.007} &\et{\textbf{2.971}}{.227}\\\midrule

\multirow{6}{*}{\makecell[l]{Human\\ML3D++}} & T2M-GPT~\cite{zhang2023generating} & \et{0.357}{.002} & \et{0.509}{.001} & \et{0.590}{.001} & \et{0.651}{.006} & \et{4.384}{.006} & \et{1.645}{.053} \\
~& T2M-GPT* & \et{0.362}{.002} & \et{0.507}{.001} & \et{0.593}{.001} & \et{0.335}{.005} & \et{4.247}{.006}  &  \et{1.984}{.051} \\
~&    MoMask (base) $^\S$ & \et{0.375}{.002} & \et{0.534}{.001} & \et{0.628}{.002} & \et{0.557}{.007} & \et{4.022}{.007}&\et{1.654}{.061}\\
~&    MoMask$^\S$~\cite{guo2024momask} & \et{0.385}{.002} & \et{0.546}{.002} & \et{{0.640}}{.002} & \et{0.380}{.005} & \et{{3.935}}{.006} &\et{1.693}{.060}\\
\cline{2-8}
~&    \modelname~(base) & \et{\underline{0.394}}{.002} & \et{\underline{0.551}}{.001} & \et{\underline{0.641}}{.001} & \et{\underline{0.189}}{.003} & \et{\underline{3.873}}{.003} &\et{\underline{2.461}}{.107}\\
~&    \textbf{\modelname~} & \et{\textbf{0.401}}{.001} & \et{\textbf{0.558}}{.002} & \et{\textbf{0.647}}{.002} & \et{\textbf{0.141}}{.002} & \et{\textbf{3.814}}{.012} &\et{\textbf{2.538}}{.114}\\\midrule

\multirow{6}{*}{\makecell[l]{KIT\\ML++}} & T2M-GPT~\cite{zhang2023generating} & \et{0.289}{.003} & \et{0.438}{.004} & \et{0.529}{.003} & \et{0.516}{.014} & \et{4.669}{.011}  & \et{1.813}{.056} \\
~&    MoMask (base) $^\S$ & \et{0.296}{.004} & \et{0.452}{.004} & \et{0.550}{.003} & \et{0.508}{.010} & \et{4.454}{.012}  &\et{1.594}{.046}\\
~&    MoMask$^\S$~\cite{guo2024momask} & \et{0.305}{.004} & \et{\textbf{0.461}}{.004} & \et{\underline{0.555}}{.003} & \et{\underline{0.425}}{.016} & \et{\underline{4.403}}{.013}  &\et{1.716}{.054}\\
\cline{2-8}
~&  \modelname~(base) & \et{\underline{0.305}}{.003} & \et{{0.456}}{.003} & \et{{0.546}}{.003} & \et{{0.545}}{.012} & \et{4.460}{.011}  &\et{\underline{2.411}}{.101}\\
~&    \textbf{\modelname~} & \et{\textbf{0.309}}{.003} & \et{\underline{0.460}}{.003} & \et{\textbf{0.557}}{.004} & \et{\textbf{0.230}}{.008} & \et{\textbf{4.388}}{.009} &\et{\textbf{2.524}}{.112}\\
\bottomrule
    \end{tabular}
    }\caption{Comparing our \textbf{\modelname} with baselines on multiple datasets. \textbf{Bold} indicates the best result, and \underline{underlined} indicates the second best result. $^\S$ indicates using ground-truth motion length as extra information. \textbf{\modelname~(base)} means using VQ-VAE without flow matching. \textbf{MoMask~(Base)} and \textbf{BAMM~(Base)} refers to using residual VQ-VAE but without residual Transformer. We implement a variant version of the original T2M-GPT, denoted as \textbf{T2M-GPT*}, which only differs from \modelname~(base) in that there is no progressive planning. MoMask and T2M-GPT are retrained by their source code on the HumanML3D++ and KIT-ML++ datasets.}\vspace{-10pt}
    \label{tab:res_ml3d}
\end{table*}

\subsection{Experimental Setup}
\noindent\textbf{Datasets.} 
We conduct evaluations on four datasets, with the statistics shown in Table~\ref{tab:data_stat}. \textbf{HumanML3D}~\cite{guo2022generating} is formed by text annotation of the HumanAct12~\cite{guo2020action2motion} and AMASS~\cite{mahmood2019amass} datasets, containing 14,616 motions and 44,970 text descriptions at 20 FPS with an average duration of 7.1 seconds.
\textbf{KIT-ML}~\cite{mandery2015kit} is an early and limited-scale dataset (3,911 motions, 6,278 texts) captured at 12.5 FPS from the sources of KIT and CMU~\cite{cmumocap}.
To further verify the ability to generate complex and long motions,
we introduce \textbf{HumanML3D++} and \textbf{KIT-ML++}, two larger-scale long motions datasets.
We follow previous work~\cite{li2024infinite} to merge 2-5 motions to long motions, generating seamless sequences of up to 50s duration (5 times the original maximum). We use GPT-4 to merge texts. We preserve the original data splits to ensure fair benchmarking. We sample 100 motion-text pairs, of which 86\% of the sequences are reliable, evaluating both motion smoothness and text-motion alignment.
The newly collected datasets achieve an expansion of 8 times (127K+ motions for HumanML3D++, 24K+ motions for KIT-ML++) with an average duration of 20.1s (2.8 times) and 27.3s (4.5 times).

\noindent\textbf{Implementation Details.}
We develop the flow-enhanced tokenizer based on previous approaches~\cite{guo2024momask,lipman2022flow}, with a learning rate of 2e-4, a batch size of 256, and up to 50 training epochs. 
For efficient flow-matching inference, motion sequences are split into 64-frame clips and re-stitched after inference. The vector field is constructed using a U-Net~\cite{ronneberger2015u} with 3 downsampling and 3 upsampling blocks, 256 max channels, and group normalization (32 groups).
The HumanML3D and HumanML3D++ datasets share the same motion tokenizer. So do KIT-ML and KIT-ML++.
The LLM is implemented by LLaMA-Factory~\cite{zheng2024llamafactory} by fine-tuning TinyLLaMA (1B parameters)~\cite{zhang2024tinyllama}, with the learning rate on HumanML3D and HumanML3D++ at 5e-5 (1000 warm-up steps), and for KIT-ML and KIT-ML++, it is 1e-3 (600 warm-up steps). Training converges in 25 epochs, significantly fewer than prior works~\cite{guo2024momask} (500 epochs). 
Training takes 4.8 hours for VQ-VAE, 15 hours for flow-matching, 28.8 hours for the LLM on HumanML3D, and 149.6 hours on HumanML3D++ on a single NVIDIA H800. KIT-ML and KIT-ML++'s training is faster due to smaller scales.

\noindent\textbf{Evaluation Metrics.}
We follow the metrics from previous works~\cite{zhang2023generating,guo2024momask}: Frechet Inception Distance~(\textbf{FID}) score evaluates the quality of the generated motion by measuring the differences in the distribution between the generated and real motions. \textbf{R-Precision} and Multimodal Distance (\textbf{MM-Dist}) evaluate the semantic alignment between motions and texts. Multimodality (\textbf{MModality}) assesses the diversity of motions generated from the same text.

\noindent\textbf{Baselines.}
We compare \modelname~with diffusion methods and token-based methods. For \textbf{diffusion-based} baselines, we compare with \textbf{MDM}~\cite{tevet2023human}, \textbf{MotionDiffuse}~\cite{zhang2022motiondiffuse}, \textbf{ReMoDiffuse}~\cite{zhang2023remodiffuse}, and
\textbf{MFM}~\cite{hu2023motion};
for \textbf{token-based} baselines, we compare with \textbf{T2M-GPT}~\cite{zhang2023generating}, \textbf{MotionGPT}~\cite{jiang2023motiongpt}, \textbf{MotionLLM}~\cite{wu2025motionagent}, \textbf{MoMask}~\cite{guo2024momask}, and \textbf{BAMM}~\cite{pinyoanuntapong2024bamm}.
The MoMask and BAMM without residual learning are also our baselines, denoted as \textbf{MoMask~(Base)} and \textbf{BAMM~(Base)}.
To validate the effectiveness of the proposed progressive planning method, we implement a variant of T2M-GPT, referred to as \textbf{T2M-GPT*}, as well as TinyLLaMA as the backbone. The only difference between T2M-GPT* and \modelname~(Base) is that T2M-GPT* does not integrate the planning mechanism.

\subsection{Main Results}
As shown in Table~\ref{tab:res_ml3d}, 
our \modelname~outperforms existing methods in most metrics on three datasets including HumanML3D++, KIT-ML++, and HumanML3D datasets. Results on various metrics show its effectiveness in semantic alignment and detail preservation.

Our \modelname~achieves substantial improvements in all metrics on the HumanML3D++ dataset, especially an FID (measuring overall quality) of 0.141 (vs. 0.380 in MoMask) and an R@1 score (measuring text-motion alignment) of 40.1\% (+1.6\% over MoMask).
\modelname~scores 2.538 in MModality, also a notable improvement of 0.845 over MoMask.
On the KIT-ML++ dataset, our \modelname~also outperforms previous works on most metrics, especially an FID of 0.230 (vs. 0.425 in MoMask) and an MMdoality of 2.524 (vs. 1.716 in MoMask). 
These all indicate that our method can generate long motions effectively in terms of quality, semantic alignment, and diversity.

Comparing \modelname~to its base version, the proposed flow-enhanced method significantly improves motion quality, particularly in FID, across both datasets. Additionally, \modelname~(Base) surpasses MoMask (Base) and BAMM (Base) in most metrics, and outperforms T2M-GPT* on both datasets, which validates the effectiveness of our progressive planning method.

\modelname~also achieves new state-of-the-art results on the HumanML3D dataset, with an R@1 score of 52.6\%, surpassing BAMM by 0.5\%, and an MM-Dist of 2.884, outperforming BAMM by 0.052. For MModality, \modelname~scores 2.971, significantly higher than all prior methods, indicating superior diversity and detail motion generation.

Although our method is effective on multiple datasets, it achieves suboptimal results on the KIT-ML dataset. We attribute this to the smaller scale and lower temporal resolution (12.5fps vs. 20fps for HumanML3D), which may limit the learning of progressive planning. Notably, \modelname~still achieves a high MModality score, which shows that our method can still generate diverse motions on this dataset. More results are provided in the supplementary materials.

\subsection{Diversity Analysis}
We further investigate the relationship between motion diversity and quality. By adjusting the temperature and Top-K parameters during inference, we enable the model to generate motions with varying levels of diversity.
Figure~\ref{fig:fid_mm} compares the performance of the base versions of \modelname, and non-LLM methods BAMM and MoMask, on the HumanML3D dataset. The performance of BAMM and MoMask significantly drops when the MModality score reaches 2.0 and deteriorates further, or even crashes, when it exceeds 3. This suggests that, while both models perform well at low MModality levels, these non-LLM methods fail to generalize and generate diverse motions due to overfitting. 
\begin{wraptable}{r}{8.5cm}
    \vspace{-8pt}
    \includegraphics[width=1.0\linewidth]{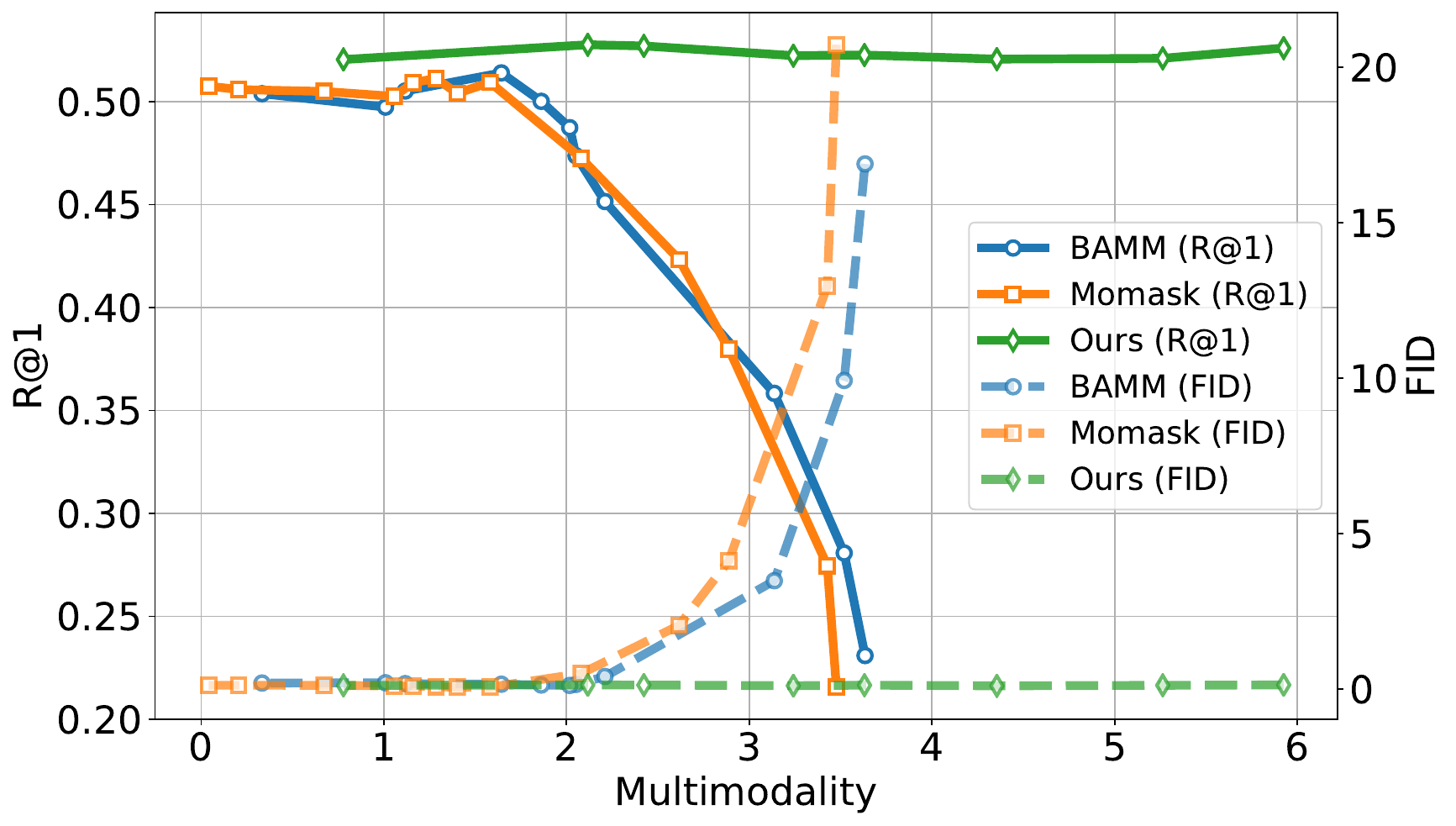}
    \caption{Exploring the relationship between MModality score and quality of generated motions. All these models are the base version. ``R@1'' refers to R-Precision Top-1.
    }
    \label{fig:fid_mm}\vspace{-13pt}
\end{wraptable}
In contrast, \modelname~maintains robust performance across a wide range of MModality levels, showing little decline in FID or R-Precision Top-1 metrics even when MModality rises to 6. 
We attribute this advantage to that \modelname~does not have the local dependency issue when generating large interval plans, thus fully exploiting LLMs' generalization power in text-to-motion tasks, which makes the generated plan richer and leads to diverse motion generation.

\begin{figure}
    \centering
    \includegraphics[width=0.9\linewidth]{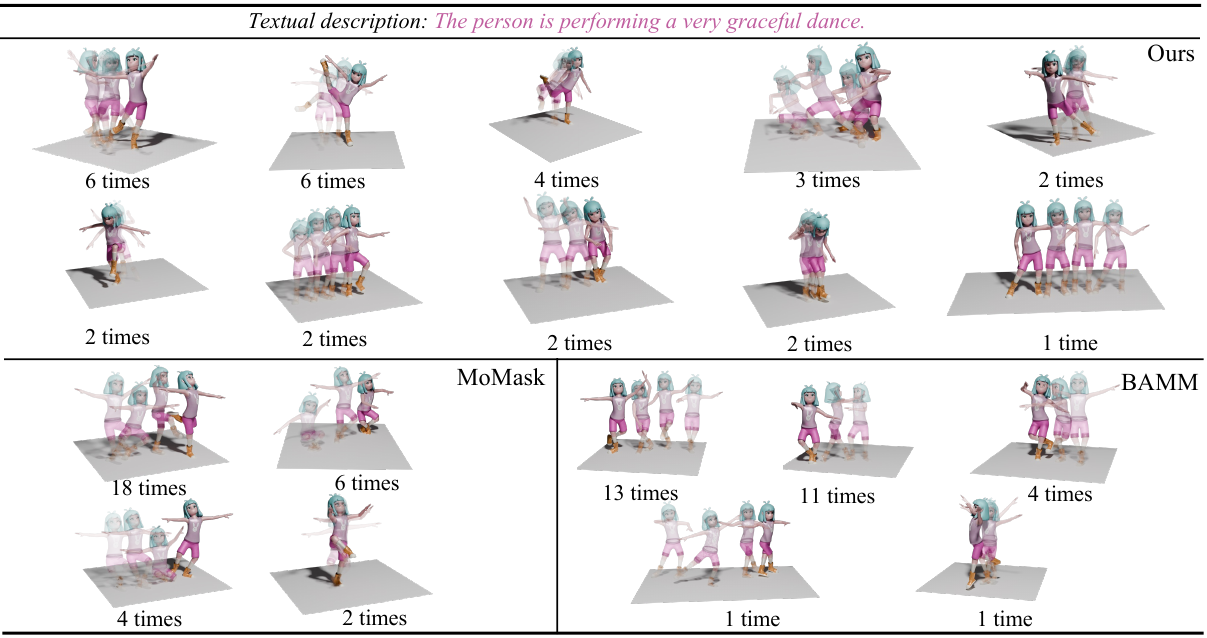}
    \caption{Different methods repeatedly generate 30 motions based on the same text. Similar motions are grouped and reported times.}\vspace{-10pt}
    \label{fig:diversity}
\end{figure}

In Figure~\ref{fig:diversity}, we show the results of different methods for repeatedly generating 30 motions based on the same text. It is obvious that the motions generated by our \modelname~are more diverse, i.e., 10 different dancing motions. Although MoMask and BAMM can generate high-quality motions, the motions they generate are relatively similar, i.e., 4 or 5 different dancing motions. This further verifies that our \modelname~can well solve the dilemma of balancing diversity and quality in existing non-LLM research.

\begin{figure*}
    \centering
    \includegraphics[width=0.9\linewidth]{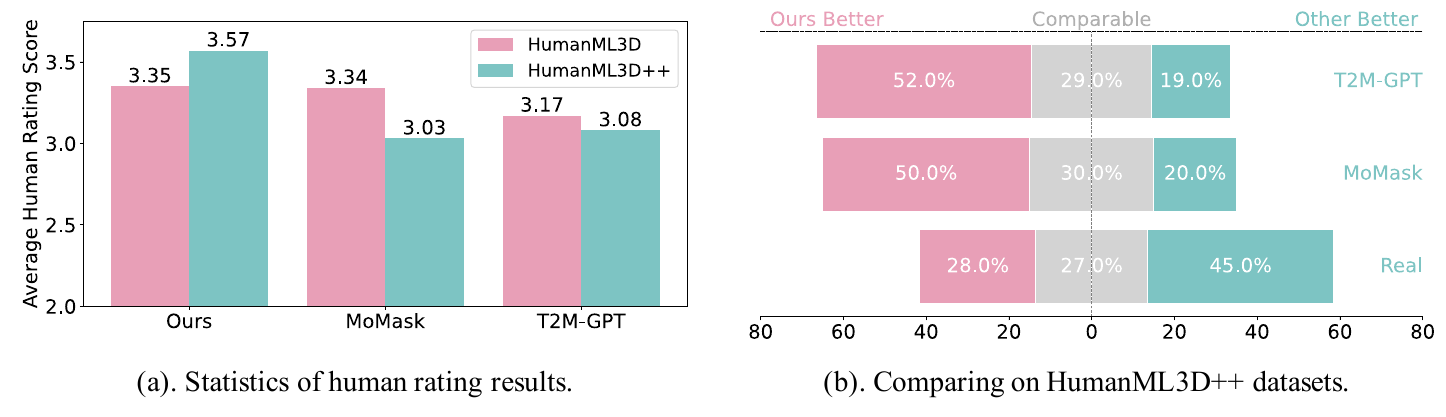}
    \caption{Case study on HumanML3D and HumanML3D++ datasets. The score range in Figure (a) ranges from 1 to 5, where 1 means poor and 5 means perfect. Figure (b) compares \modelname~and other methods~(ground-truth) to which one generates better results.}\vspace{-10pt}
    \label{fig:case_study}
\end{figure*}

\subsection{Subjective Evaluation}
We conduct a subjective evaluation by randomly selecting 200 text samples from the test sets of HumanML3D++ and HumanML3D dataset, and inviting five volunteers to rate the motions generated by different methods. 
Results are shown in Figure~\ref{fig:case_study}. On the HumanML3D and HumanML3D++ datasets, our \modelname~scores 3.35 and 3.57, respectively, significantly better than T2M-GPT and MoMask. 
Furthermore, the comparison with baselines in the HumanML3D++ dataset in Figure~\ref{fig:case_study}~(b) further confirms \modelname's advantage in generating long-sequence motions.

\subsection{Ablation Study}

\noindent\textbf{Interval of Progressive Planning.}
We investigate the impact of the plans, with the results on the HumanML3D test dataset shown in Table~\ref{tab:ablation_scale}. 
Compared to the model without plans, all the models using plans show improvements in both R@1 and MM-Dist metrics. 
\begin{wraptable}{r}{7cm}
    \centering\scalebox{0.75}{
    \begin{tabular}{cccccc}\toprule
    \multicolumn{3}{c}{Interval $T$} & \multirow{2}{*}{R@1~$\uparrow$} & \multirow{2}{*}{FID $\downarrow$} & \multirow{2}{*}{MM-Dist~$\downarrow$} \\\cline{1-3}
       8 & 4 & 2 & ~ & ~ & \\\midrule
       &  &  & \et{0.494}{.003} & \et{0.130}{.004} & \et{3.109}{.006} \\
      $\checkmark$ &  &  &  \et{0.506}{.003} & \et{0.162}{.005} & \et{3.013}{.010} \\
      & $\checkmark$  &  & \et{0.514}{.003} & \et{0.128}{.006} & \et{2.969}{.009} \\
       &  & $\checkmark$ & \et{0.508}{.002} & \et{\textbf{0.104}}{.004} & \et{3.030}{.008} \\
$\checkmark$  & $\checkmark$   &  & \et{\underline{0.518}}{.003} & \et{0.175}{.004} & \et{\textbf{2.930}}{.007} \\
    & $\checkmark$ & $\checkmark$ & \et{\textbf{0.521}}{.003} & \et{\underline{0.106}}{.004} & \et{\underline{2.937}}{.008} \\
 $\checkmark$ & $\checkmark$ & $\checkmark$ & \et{\underline{0.519}}{.002} & \et{0.119}{.006} & \et{2.973}{.010} \\
         \bottomrule
    \end{tabular}
    }\caption{Analysis of progressive planning method based on the HumanML3D test dataset. }\vspace{-10pt}
    \label{tab:ablation_scale}
\end{wraptable}
When the interval of the plan is 2, it significantly reduces FID, indicating that the model can generate more refined motions. 
When the interval of the plan is 4, R@1 improves while MM-Dist decreases, suggesting that the generated motions are more semantically aligned with the text. 
Combining these two interval plans leads to improvements both in FID and R@1, showing that the enhancements provided by different plans are cumulative. 
With an interval of 8, although R@1 increases, FID also rises, indicating that a larger interval plan may reduce the fineness of the generated motions.
When all three intervals are used, there is no further performance improvement. 
This shows that our selection of intervals 4 and 2 is sufficient.

\noindent\textbf{Granularity of Motion Tokenization.} 
Table~\ref{tab:sub1} shows codebook size and downsampling rate effects.
As codebook size increases, both reconstruction and generation performance improve. 
Notably, with downsampling rate 4, LLM performance improves marginally with larger codebooks, while a downsampling rate of 2 shows substantial gains, surpassing rate 4 at a 4096 codebook size. This indicates that rate 4 requires smaller codebooks as it captures less detail. We therefore select a codebook size of 4096 with a downsampling rate of 2.

\begin{table}[t]
\centering
\begin{subtable}[t]{0.48\textwidth} % [t]顶部对齐
\centering
\scalebox{0.68}{
    \begin{tabular}{cccccc}\toprule
    \multirow{2}{*}{VQ-VAE} & \multicolumn{2}{c}{Reconstruction} & & \multicolumn{2}{c}{Generation}  \\\cline{2-3} \cline{5-6}
 & FID $\downarrow$ & MPJPE $\downarrow$ &  & FID $\downarrow$ & MM-Dist $\downarrow$ \\\midrule
512, 4 & \et{0.088}{.002} & 67.6 &  & \et{0.185}{.007} & \et{2.987}{.009} \\
512, 2 & \et{0.146}{.001} & 59.0 &  & \et{0.244}{.007} & \et{3.045}{.009} \\
1024, 4 & \et{0.075}{.001} & 66.8 &  & \et{0.164}{.005} & \et{2.969}{.008} \\
1024, 2 & \et{0.119}{.000} & 59.0 &  & \et{0.221}{.007} & \et{2.987}{.008} \\
2048, 4 & \et{\underline{0.061}}{.001} & 66.4 &  & \et{0.147}{.004} & \et{2.974}{.007} \\
2048, 2 & \et{0.074}{.001} & \textbf{54.4} &  & \et{0.142}{.005} & \et{3.002}{.007} \\
4096, 4 & \et{\textbf{0.057}}{.000} & 66.7 &  & \et{\underline{0.130}}{.005} & \et{\underline{2.967}}{.006} \\
4096, 2 & \et{0.066}{.000} & \underline{56.7} &  & \et{\textbf{0.106}}{.004} & \et{\textbf{2.937}}{.008} \\
         \bottomrule
    \end{tabular}
}\caption{Impact of granularity of motion tokenization.}
\label{tab:sub1}
\end{subtable}
\hfill % 自动填充间距
\begin{subtable}[t]{0.48\textwidth}
\centering
\scalebox{0.64}{
    \begin{tabular}{lcccccc}\toprule
    VQ-VAE & Size & Model & R@1 $\uparrow$ & FID $\downarrow$ & MM-Dist $\downarrow$ \\\midrule
    Residual & 512, 4  & Base & \et{0.375}{.002} & \et{0.431}{.005} & \et{4.031}{.004} \\
    Residual & 1024, 4 & Base & \et{0.379}{.002} & \et{0.497}{.006} & \et{4.012}{.004} \\
    Residual & 2048, 2 & Base & \et{\underline{0.385}}{.002} & \et{\underline{0.334}}{.004} & \et{\underline{3.970}}{.005} \\
    Residual & 4096, 2 & Base & \et{0.378}{.002} & \et{0.480}{.006} & \et{4.077}{.006} \\
    \textbf{Flow} & 4096, 2 & Base & \et{\textbf{0.394}}{.002} & \et{\textbf{0.189}}{.003} & \et{\textbf{3.873}}{.003} \\\midrule
    Residual & 512, 4 & Full & \et{0.392}{.002} & \et{0.242}{.004} & \et{3.858}{.004} \\
    Residual & 1024, 4 & Full & \et{0.391}{.002} & \et{\underline{0.165}}{.004} & \et{\underline{3.849}}{.004} \\
    Residual & 2048, 2 & Full & \et{\underline{0.396}}{.002} & \et{0.188}{.003} & \et{3.857}{.005} \\
    Residual & 4096, 2 & Full & \et{0.390}{.002} & \et{0.208}{.004} & \et{3.909}{.005} \\
    \textbf{Flow} & 4096, 2 & Full & \et{\textbf{0.401}}{.001} & \et{\textbf{0.141}}{.002} & \et{\textbf{3.814}}{.012} \\
    \bottomrule
    \end{tabular}
    }\caption{Comparison of residual and flow-enhanced.}
\label{tab:sub2}
\end{subtable}
\caption{(a). The impact of granularity of motion tokenization to the \modelname~(Base) on the HumanML3D test dataset. ``4096, 2'' refers to the size of the codebook is 4096, and the downsampling rate is 2. (b). Ablation analysis of flow-enhanced VQ-VAE and residual VQ-VAE on HumanML3D++ dataset. ``4096, 2'' refers to the size of the codebook and the downsampling rate is 4096 and 2. ``base'' refers to not using residual Transformer or flow-enhanced method.}\vspace{-10pt}
\label{tab:ab_vqvae}
\end{table}

\noindent\textbf{Flow-enhanced VQ-VAE vs. Residual VQ-VAE.}
We compare the flow-enhanced VQ-VAE with the previous residual VQ-VAE, which is proposed to reconstruct motion lost due to vector quantization methods in MoMask~\cite{guo2024momask}. Following MoMask's setting, we train a 6-layer residual VQ-VAE, using its first-layer tokens for the \modelname~(Base) and all six layers' tokens for the residual Transformer. 
As shown in Table~\ref{tab:sub2}, our flow-enhanced VQ-VAE outperforms residual VQ-VAE, regardless of codebook sizes and downsampling rates. Our analysis suggests that the 6-layer residual VQ-VAE will compromise the quality of the base token, thereby weakening the following base generative model. 
The reconstruction performance of VQVAEs is reported in the supplementary materials.

\begin{figure}[t]
\centering
\begin{subfigure}[b]{0.48\textwidth}
    \includegraphics[width=\linewidth]{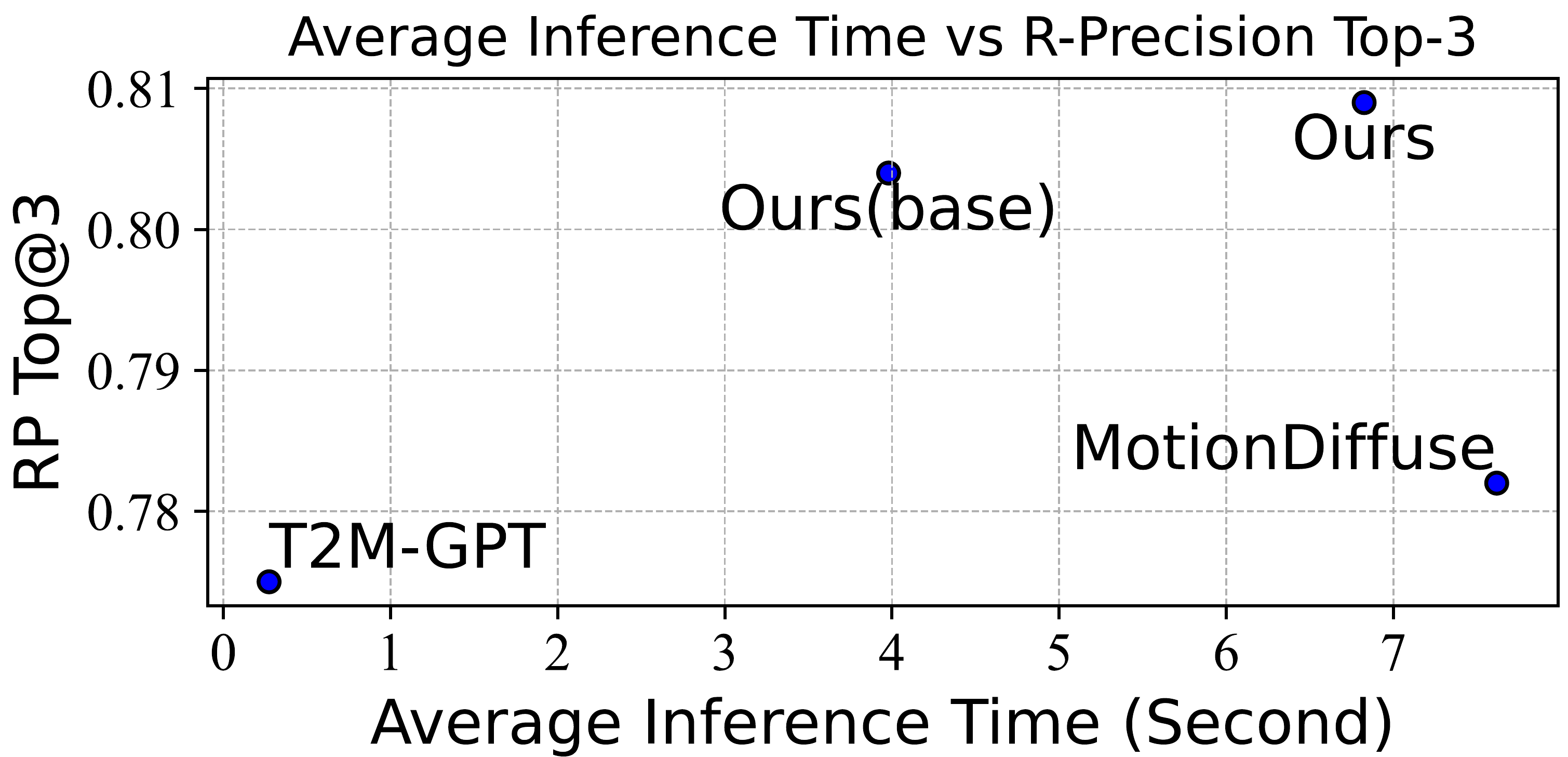}
    \caption{Comparisons on RP Top-3 and Inference Cost}
    \label{fig:time_cost}
\end{subfigure}
\hfill % 添加水平间距
\begin{subfigure}[b]{0.48\textwidth}
    \includegraphics[width=\linewidth]{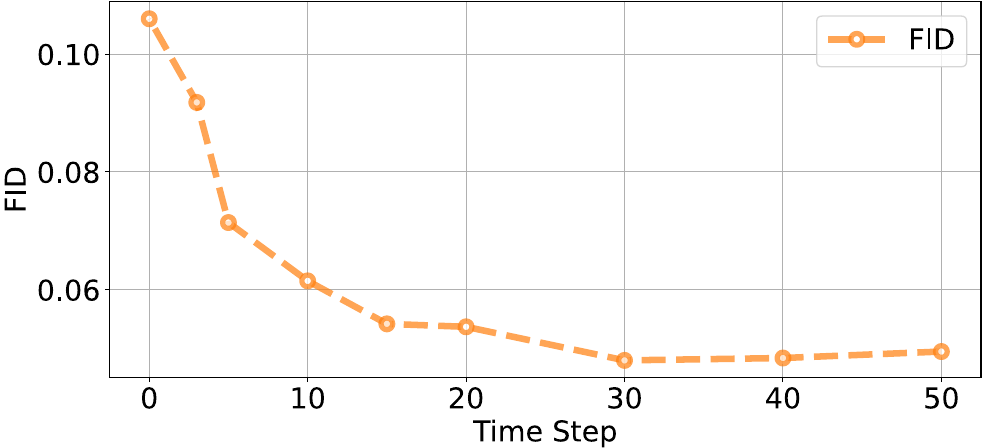}
    \caption{Inference Step of Flow-Matching}
    \label{fig:flow_time}
\end{subfigure}
\caption{(a). Comparisons of average inference time cost. (b). Exploring the inference step of the flow-matching enhanced method on the HumanML3D test dataset.}\vspace{-10pt}
\label{fig:combined}
\end{figure}

\noindent\textbf{Analysis of Time Cost.}
Figure~\ref{fig:time_cost} report the time cost of different methods, which is statistics from 100 cases.
\modelname~is slower than T2M-GPT but still faster than diffusion-based MotionDiffuse. Despite the speed gap with T2M-GPT, our approach achieves state-of-the-art motion quality (RP@3 0.81 vs 0.68)
% Our method achieves a quality-speed balance.
This shows that our method achieves a quality-speed balance.

\noindent\textbf{Time Steps of Flow Matching.}
Figure~\ref{fig:flow_time} reports the impact of the inference steps of the flow matching. As the steps increase, the FID significantly decreases and converges at around 30 steps. 

%% file: sec/conclusion.tex
\section{Conclusion}
\label{sec:conclusion}

In this paper, we address what limits the ability of LLM in text-to-motion generation tasks. We identify the granularity of motion tokenization as a critical bottleneck. That is, fine-grained tokens lead to severe local dependencies, while coarse-grained motion tokens lose motion details.
To address this issue, we propose \modelname, an LLM-based framework integrating progressive planning and flow-enhanced fine-grained motion tokenization.
Extensive experiments demonstrate that \modelname~not only generates precise and diverse human motions but also outperforms the state-of-the-art method MoMask in both human and automated evaluations on short and long sequence datasets.
What's more, \modelname~resolves the diversity-quality dilemma in existing non-LLM approaches. which further verifies the necessity of exploiting the potential of LLM for text-to-motion tasks.
In the future, we will explore more flexible planning, such as manually selected keyframes. In addition, we will also explore how to extend \modelname~to generate motion with expressions and hand movements.

%% file: sec/appendix.tex
\begin{table}[]
    \centering
    \scalebox{0.81}{
    \begin{tabular}{l l c c c c c c}
    \toprule
\multirow{2}{*}{Datasets} & \multirow{2}{*}{Methods}  & \multicolumn{3}{c}{R-Precision $\uparrow$} & \multirow{2}{*}{FID $\downarrow$} & \multirow{2}{*}{MM-Dist $\downarrow$} & \multirow{2}{*}{MModality $\uparrow$}\\
    \cline{3-5}
    ~ & ~ & Top-1 & Top-2 & Top-3 \\\midrule
\multirow{9}{*}{\makecell[l]{KIT-ML}} & Real Motion & \et{0.424}{.005} & \et{0.649}{.006} & \et{0.779}{.006} & \et{0.031}{.004} & \et{2.788}{.012} & -  \\
\cline{2-8}
~ &    MDM$^\S$ & - & - & \et{0.396}{.004} & \et{0.497}{.021}  & \et{1.907}{.214}  \\ 
~ &    MotionDiffuse$^\S$ & \et{0.417}{.004} & \et{0.621}{.004} & \et{0.739}{.004} & \et{1.954}{.062} & \et{2.958}{.005}& \et{0.730}{.013}  \\ 

~ &    ReMoDiffuse$^\S$ & \et{{0.427}}{.014} & \et{{0.641}}{.004} & \et{{0.765}}{.055} & \et{\textbf{0.155}}{.006} & \et{\underline{2.814}}{.012}  &\et{1.239}{.028}\\
\cline{2-8}
~ &    T2M-GPT & \et{0.416}{.006} & \et{0.627}{.006} & \et{0.745}{.006} & \et{0.514}{.029} & \et{3.007}{.023} &  \et{1.570}{.039} \\
~ &    MotionGPT & \et{0.366}{.005} & \et{0.558}{.004} & \et{0.680}{.005} & \et{0.510}{.016} & \et{3.527}{.021} &  \et{2.328}{.117} \\

~ &    MoMask (base) $^\S$ & \et{0.415}{.010} & \et{0.634}{.011} & \et{0.760}{.005} & \et{0.372}{.020} & \et{2.931}{.041} &\et{1.097}{.054}\\
~ &    MoMask$^\S$ & \et{\underline{0.433}}{.007} & \et{\underline{0.656}}{.005} & \et{\underline{0.781}}{.005} & \et{0.204}{.011} & \et{\underline{2.799}}{.022} &\et{1.131}{.043}\\
~&    BAMM$^\S$ & \et{\textbf{0.436}}{.007} & \et{\textbf{0.660}}{.006} & \et{\textbf{0.791}}{.005} & \et{{0.200}}{.011} & \et{\textbf{2.714}}{.016} &\et{1.517}{.058}\\
\cline{2-8}
~ &    \modelname~(base) & \et{0.422}{.007} & \et{0.625}{.007} & \et{0.742}{.007} & \et{0.359}{.012} & \et{2.962}{.019} & \et{\textbf{2.412}}{.131}\\
~ &    \textbf{\modelname} & \et{0.422}{.005} & \et{0.631}{.006} & \et{0.755}{.005} & \et{\underline{0.193}}{.007} & \et{2.964}{.015} & \et{\underline{2.391}}{.109} \\
\bottomrule
    \end{tabular}
    }\caption{Comparing our \textbf{\modelname} with baselines on the \textbf{KIT-ML} test dataset. \textbf{Bold} indicates the best result, and \underline{underlined} indicates the second best result. $^\S$ indicates using ground-truth motion length as extra information. \textbf{\modelname~(base)} means not including flow-matching refinement. \textbf{MoMask~(Base)} refers to using residual VQ-VAE but without residual Transformer. }
    \label{tab:res_kit}
\end{table}

\section{Evaluation on KIT-ML dataset}
The evaluation results on the KIT-ML dataset are shown in Table~\ref{tab:res_kit}. It achieves suboptimal results on the KIT-ML dataset. We attribute this to the smaller scale and lower temporal resolution (12.5fps vs. 20fps for HumanML3D), which may limit the learning of progressive planning. Notably, \modelname~still achieves a high MModality score, which shows that our method can still generate diverse motions on this dataset. More results are provided in the supplementary materials.

\begin{table*}[t]
    \centering
    \scalebox{0.85}{
    \begin{tabular}{l c c c c c c c c c}
    \toprule
\multirow{2}{*}{VQ-VAE} & \multirow{2}{*}{Size}  & \multirow{2}{*}{Model} & \multicolumn{3}{c}{R-Precision $\uparrow$} & \multirow{2}{*}{FID $\downarrow$} & \multirow{2}{*}{MM-Dist $\downarrow$} & \multirow{2}{*}{MPJEG $\downarrow$}\\
    \cline{4-6}
    ~ & ~ & ~ & Top-1 & Top-2 & Top-3 \\\midrule
Residual & 512, 4 & Base & \et{0.474}{.003} & \et{0.666}{.002} & \et{0.766}{.002} & \et{0.227}{.002} & \et{3.220}{.009}  & \et{71.3}{.100} \\
Residual & 1024, 4  & Base & \et{0.485}{.003} & \et{0.676}{.002} & \et{0.773}{.002} & \et{0.172}{.001} & \et{3.162}{.007}  & \et{67.0}{.100} \\
Residual & 2048, 2  & Base & \et{0.493}{.002} & \et{0.684}{.003} & \et{0.780}{.002} & \et{0.163}{.000} & \et{3.105}{.007}  & \et{64.2}{.100} \\
Residual & 4096, 2  & Base & \et{0.493}{.003} & \et{0.685}{.003} & \et{0.782}{.002} & \et{0.173}{.001} & \et{3.106}{.009}  & \et{65.3}{.200} \\
Flow & 4096, 2 & Base &\et{\textbf{0.502}}{.002} & \et{\textbf{0.695}}{.003} & \et{\textbf{0.788}}{.002} & \et{\textbf{0.066}}{.000} & \et{\textbf{3.020}}{.006} & \et{\textbf{56.8}}{.200} \\\midrule
Residual & 512, 4 & Full & \et{0.507}{.002} & \et{0.699}{.003} & \et{0.794}{.002} & \et{0.033}{.000} & \et{3.016}{.006} & \et{40.3}{.200} \\
Residual & 1024, 4  & Full& \et{0.506}{.002} & \et{0.698}{.002} & \et{0.794}{.002} & \et{0.022}{.000} & \et{3.019}{.008}  & \et{37.4}{.200} \\
Residual & 2048, 2  & Full& \et{0.511}{.003} & \et{0.703}{.002} & \et{\textbf{0.797}}{.002} & \et{0.022}{.000} & \et{2.996}{.008} & \et{\textbf{31.6}}{.200} \\
Residual & 4096, 2  & Full & \et{\textbf{0.512}}{.002} & \et{\textbf{0.703}}{.003} & \et{0.796}{.002} & \et{0.022}{.000} & \et{\textbf{2.993}}{.009}  & \et{31.7}{.300} \\
Flow & 4096, 2 & Full & \et{0.506}{.003} & \et{0.698}{.002} & \et{0.792}{.002} & \et{\textbf{0.014}}{.000} & \et{2.995}{.007}  & \et{48.1}{.100} \\
\bottomrule
    \end{tabular}
    }\caption{Reconstruction of our flow-enhanced VQ-VAE and residual VQ-VAE on the \textbf{HumanML3D} test dataset. ``base'' for ``residual'' refers to the motion is reconstructed by residual VQ-VAE without residual tokens; ``base'' for ``Flow'' refers to our VQ-VAE without flow matching method; ``4096, 2'' refers to the size of the codebook is 4096, and the downsampling rate is 2.}
    \label{tab:vqvae_recon}
\end{table*}

\section{Motion Reconstruction Results}
We show the motion reconstruction results of flow-enhanced VQ-VAE and residual VQ-VAE on the HumanML3D dataset in Table~\ref{tab:vqvae_recon}. 
Although the 6-layer stacked residual VQ-VAE can achieve better reconstruction results overall, the reconstructed quality from the base token is not suboptimal (corresponding to the ``base'' in Table~\ref{tab:vqvae_recon}). This is because residual VQ-VAE disperses the motion details into the residual tokens of each layer, making it difficult for the base tokens to retain detailed information. This will affect the subsequent motion generation model, as shown in Table 5 in the main text, resulting in suboptimal results.

%% file: neurips_2025.bbl
\begin{thebibliography}{10}

\bibitem{hu2023motion}
Motion flow matching for human motion synthesis and editing.
\newblock {\em arXiv preprint arXiv:2312.08895}, 2023.

\bibitem{achiam2023gpt}
Josh Achiam, Steven Adler, Sandhini Agarwal, Lama Ahmad, Ilge Akkaya, Florencia~Leoni Aleman, Diogo Almeida, Janko Altenschmidt, Sam Altman, Shyamal Anadkat, et~al.
\newblock Gpt-4 technical report.
\newblock {\em arXiv preprint arXiv:2303.08774}, 2023.

\bibitem{alayrac2022flamingo}
Jean-Baptiste Alayrac, Jeff Donahue, Pauline Luc, Antoine Miech, Iain Barr, Yana Hasson, Karel Lenc, Arthur Mensch, Katherine Millican, Malcolm Reynolds, et~al.
\newblock Flamingo: a visual language model for few-shot learning.
\newblock {\em Advances in neural information processing systems}, 35:23716--23736, 2022.

\bibitem{ao2023gesturediffuclip}
Tenglong Ao, Zeyi Zhang, and Libin Liu.
\newblock Gesturediffuclip: Gesture diffusion model with clip latents.
\newblock {\em ACM Transactions on Graphics (TOG)}, 42(4):1--18, 2023.

\bibitem{brown2020language}
Tom Brown, Benjamin Mann, Nick Ryder, Melanie Subbiah, Jared~D Kaplan, Prafulla Dhariwal, Arvind Neelakantan, Pranav Shyam, Girish Sastry, Amanda Askell, et~al.
\newblock Language models are few-shot learners.
\newblock {\em Advances in neural information processing systems}, 33:1877--1901, 2020.

\bibitem{cao2020long}
Zhe Cao, Hang Gao, Karttikeya Mangalam, Qi-Zhi Cai, Minh Vo, and Jitendra Malik.
\newblock Long-term human motion prediction with scene context.
\newblock In {\em Computer Vision--ECCV 2020: 16th European Conference, Glasgow, UK, August 23--28, 2020, Proceedings, Part I 16}, pages 387--404. Springer, 2020.

\bibitem{cmumocap}
{Carnegie Mellon University}.
\newblock Cmu graphics lab motion capture database, 2003.

\bibitem{chang2017matterport3d}
Angel Chang, Angela Dai, Thomas Funkhouser, Maciej Halber, Matthias Niessner, Manolis Savva, Shuran Song, Andy Zeng, and Yinda Zhang.
\newblock Matterport3d: Learning from rgb-d data in indoor environments.
\newblock {\em arXiv preprint arXiv:1709.06158}, 2017.

\bibitem{chen2023executing}
Xin Chen, Biao Jiang, Wen Liu, Zilong Huang, Bin Fu, Tao Chen, and Gang Yu.
\newblock Executing your commands via motion diffusion in latent space.
\newblock In {\em Proceedings of the IEEE/CVF Conference on Computer Vision and Pattern Recognition}, pages 18000--18010, 2023.

\bibitem{degardin2022generative}
Bruno Degardin, Joao Neves, Vasco Lopes, Joao Brito, Ehsan Yaghoubi, and Hugo Proen{\c{c}}a.
\newblock Generative adversarial graph convolutional networks for human action synthesis.
\newblock In {\em Proceedings of the IEEE/CVF Winter Conference on Applications of Computer Vision}, pages 1150--1159, 2022.

\bibitem{ginosar2019learning}
Shiry Ginosar, Amir Bar, Gefen Kohavi, Caroline Chan, Andrew Owens, and Jitendra Malik.
\newblock Learning individual styles of conversational gesture.
\newblock In {\em Proceedings of the IEEE/CVF Conference on Computer Vision and Pattern Recognition}, pages 3497--3506, 2019.

\bibitem{goodfellow2014generative}
Ian Goodfellow, Jean Pouget-Abadie, Mehdi Mirza, Bing Xu, David Warde-Farley, Sherjil Ozair, Aaron Courville, and Yoshua Bengio.
\newblock Generative adversarial nets.
\newblock {\em Advances in neural information processing systems}, 27, 2014.

\bibitem{guo2024momask}
Chuan Guo, Yuxuan Mu, Muhammad~Gohar Javed, Sen Wang, and Li~Cheng.
\newblock Momask: Generative masked modeling of 3d human motions.
\newblock In {\em Proceedings of the IEEE/CVF Conference on Computer Vision and Pattern Recognition}, pages 1900--1910, 2024.

\bibitem{guo2022generating}
Chuan Guo, Shihao Zou, Xinxin Zuo, Sen Wang, Wei Ji, Xingyu Li, and Li~Cheng.
\newblock Generating diverse and natural 3d human motions from text.
\newblock In {\em Proceedings of the IEEE/CVF Conference on Computer Vision and Pattern Recognition}, pages 5152--5161, 2022.

\bibitem{guo2020action2motion}
Chuan Guo, Xinxin Zuo, Sen Wang, Shihao Zou, Qingyao Sun, Annan Deng, Minglun Gong, and Li~Cheng.
\newblock Action2motion: Conditioned generation of 3d human motions.
\newblock In {\em Proceedings of the 28th ACM International Conference on Multimedia}, pages 2021--2029, 2020.

\bibitem{hassan2021stochastic}
Mohamed Hassan, Duygu Ceylan, Ruben Villegas, Jun Saito, Jimei Yang, Yi~Zhou, and Michael~J Black.
\newblock Stochastic scene-aware motion prediction.
\newblock In {\em Proceedings of the IEEE/CVF International Conference on Computer Vision}, pages 11374--11384, 2021.

\bibitem{ho2020denoising}
Jonathan Ho, Ajay Jain, and Pieter Abbeel.
\newblock Denoising diffusion probabilistic models.
\newblock {\em Advances in neural information processing systems}, 33:6840--6851, 2020.

\bibitem{hong2022avatarclip}
Fangzhou Hong, Mingyuan Zhang, Liang Pan, Zhongang Cai, Lei Yang, and Ziwei Liu.
\newblock Avatarclip: zero-shot text-driven generation and animation of 3d avatars.
\newblock {\em ACM Transactions on Graphics (TOG)}, 41(4):1--19, 2022.

\bibitem{jiang2023motiongpt}
Biao Jiang, Xin Chen, Wen Liu, Jingyi Yu, Gang Yu, and Tao Chen.
\newblock Motiongpt: Human motion as a foreign language.
\newblock {\em Advances in Neural Information Processing Systems}, 36:20067--20079, 2023.

\bibitem{kingma2013auto}
Diederik~P Kingma.
\newblock Auto-encoding variational bayes.
\newblock {\em arXiv preprint arXiv:1312.6114}, 2013.

\bibitem{koh2023generating}
Jing~Yu Koh, Daniel Fried, and Russ~R Salakhutdinov.
\newblock Generating images with multimodal language models.
\newblock {\em Advances in Neural Information Processing Systems}, 36:21487--21506, 2023.

\bibitem{kojima2022large}
Takeshi Kojima, Shixiang~Shane Gu, Machel Reid, Yutaka Matsuo, and Yusuke Iwasawa.
\newblock Large language models are zero-shot reasoners.
\newblock {\em Advances in neural information processing systems}, 35:22199--22213, 2022.

\bibitem{kucherenko2019analyzing}
Taras Kucherenko, Dai Hasegawa, Gustav~Eje Henter, Naoshi Kaneko, and Hedvig Kjellstr{\"o}m.
\newblock Analyzing input and output representations for speech-driven gesture generation.
\newblock In {\em Proceedings of the 19th ACM International Conference on Intelligent Virtual Agents}, pages 97--104, 2019.

\bibitem{li2021audio2gestures}
Jing Li, Di~Kang, Wenjie Pei, Xuefei Zhe, Ying Zhang, Zhenyu He, and Linchao Bao.
\newblock Audio2gestures: Generating diverse gestures from speech audio with conditional variational autoencoders.
\newblock In {\em Proceedings of the IEEE/CVF International Conference on Computer Vision}, pages 11293--11302, 2021.

\bibitem{li2024infinite}
Mengtian Li, Chengshuo Zhai, Shengxiang Yao, Zhifeng Xie, Keyu Chen, and Yu-Gang Jiang.
\newblock Infinite motion: Extended motion generation via long text instructions.
\newblock {\em arXiv preprint arXiv:2407.08443}, 2024.

\bibitem{li2021ai}
Ruilong Li, Shan Yang, David~A Ross, and Angjoo Kanazawa.
\newblock Ai choreographer: Music conditioned 3d dance generation with aist++.
\newblock In {\em Proceedings of the IEEE/CVF International Conference on Computer Vision}, pages 13401--13412, 2021.

\bibitem{liang2024survey}
Zijing Liang, Yanjie Xu, Yifan Hong, Penghui Shang, Qi~Wang, Qiang Fu, and Ke~Liu.
\newblock A survey of multimodel large language models.
\newblock In {\em Proceedings of the 3rd International Conference on Computer, Artificial Intelligence and Control Engineering}, pages 405--409, 2024.

\bibitem{lipman2022flow}
Yaron Lipman, Ricky~TQ Chen, Heli Ben-Hamu, Maximilian Nickel, and Matt Le.
\newblock Flow matching for generative modeling.
\newblock {\em arXiv preprint arXiv:2210.02747}, 2022.

\bibitem{liu2022beat}
Haiyang Liu, Zihao Zhu, Naoya Iwamoto, Yichen Peng, Zhengqing Li, You Zhou, Elif Bozkurt, and Bo~Zheng.
\newblock Beat: A large-scale semantic and emotional multi-modal dataset for conversational gestures synthesis.
\newblock In {\em European conference on computer vision}, pages 612--630. Springer, 2022.

\bibitem{liu2023visual}
Haotian Liu, Chunyuan Li, Qingyang Wu, and Yong~Jae Lee.
\newblock Visual instruction tuning.
\newblock {\em Advances in neural information processing systems}, 36:34892--34916, 2023.

\bibitem{mahmood2019amass}
Naureen Mahmood, Nima Ghorbani, Nikolaus~F Troje, Gerard Pons-Moll, and Michael~J Black.
\newblock Amass: Archive of motion capture as surface shapes.
\newblock In {\em Proceedings of the IEEE/CVF international conference on computer vision}, pages 5442--5451, 2019.

\bibitem{mandery2015kit}
Christian Mandery, {\"O}mer Terlemez, Martin Do, Nikolaus Vahrenkamp, and Tamim Asfour.
\newblock The kit whole-body human motion database.
\newblock In {\em 2015 International Conference on Advanced Robotics (ICAR)}, pages 329--336. IEEE, 2015.

\bibitem{men2022gan}
Qianhui Men, Hubert~PH Shum, Edmond~SL Ho, and Howard Leung.
\newblock Gan-based reactive motion synthesis with class-aware discriminators for human--human interaction.
\newblock {\em Computers \& Graphics}, 102:634--645, 2022.

\bibitem{petrovich2021action}
Mathis Petrovich, Michael~J Black, and G{\"u}l Varol.
\newblock Action-conditioned 3d human motion synthesis with transformer vae.
\newblock In {\em Proceedings of the IEEE/CVF International Conference on Computer Vision}, pages 10985--10995, 2021.

\bibitem{pinyoanuntapong2024bamm}
Ekkasit Pinyoanuntapong, Muhammad~Usama Saleem, Pu~Wang, Minwoo Lee, Srijan Das, and Chen Chen.
\newblock Bamm: Bidirectional autoregressive motion model.
\newblock In {\em Computer Vision -- ECCV 2024}, 2024.

\bibitem{punnakkal2021babel}
Abhinanda~R Punnakkal, Arjun Chandrasekaran, Nikos Athanasiou, Alejandra Quiros-Ramirez, and Michael~J Black.
\newblock Babel: Bodies, action and behavior with english labels.
\newblock In {\em Proceedings of the IEEE/CVF Conference on Computer Vision and Pattern Recognition}, pages 722--731, 2021.

\bibitem{radford2021learning}
Alec Radford, Jong~Wook Kim, Chris Hallacy, Aditya Ramesh, Gabriel Goh, Sandhini Agarwal, Girish Sastry, Amanda Askell, Pamela Mishkin, Jack Clark, et~al.
\newblock Learning transferable visual models from natural language supervision.
\newblock In {\em International conference on machine learning}, pages 8748--8763. PmLR, 2021.

\bibitem{ronneberger2015u}
Olaf Ronneberger, Philipp Fischer, and Thomas Brox.
\newblock U-net: Convolutional networks for biomedical image segmentation.
\newblock In {\em Medical image computing and computer-assisted intervention--MICCAI 2015: 18th international conference, Munich, Germany, October 5-9, 2015, proceedings, part III 18}, pages 234--241. Springer, 2015.

\bibitem{shafir2024human}
Yoni Shafir, Guy Tevet, Roy Kapon, and Amit~Haim Bermano.
\newblock Human motion diffusion as a generative prior.
\newblock In {\em The Twelfth International Conference on Learning Representations}, 2024.

\bibitem{song2020score}
Yang Song, Jascha Sohl-Dickstein, Diederik~P Kingma, Abhishek Kumar, Stefano Ermon, and Ben Poole.
\newblock Score-based generative modeling through stochastic differential equations.
\newblock {\em arXiv preprint arXiv:2011.13456}, 2020.

\bibitem{taheri2022goal}
Omid Taheri, Vasileios Choutas, Michael~J Black, and Dimitrios Tzionas.
\newblock Goal: Generating 4d whole-body motion for hand-object grasping.
\newblock In {\em Proceedings of the IEEE/CVF Conference on Computer Vision and Pattern Recognition}, pages 13263--13273, 2022.

\bibitem{tevet2022motionclip}
Guy Tevet, Brian Gordon, Amir Hertz, Amit~H Bermano, and Daniel Cohen-Or.
\newblock Motionclip: Exposing human motion generation to clip space.
\newblock In {\em European Conference on Computer Vision}, pages 358--374. Springer, 2022.

\bibitem{tevet2023human}
Guy Tevet, Sigal Raab, Brian Gordon, Yoni Shafir, Daniel Cohen-or, and Amit~Haim Bermano.
\newblock Human motion diffusion model.
\newblock In {\em The Eleventh International Conference on Learning Representations}, 2023.

\bibitem{van2017neural}
Aaron Van Den~Oord, Oriol Vinyals, et~al.
\newblock Neural discrete representation learning.
\newblock {\em Advances in neural information processing systems}, 30, 2017.

\bibitem{wang2021scene}
Jingbo Wang, Sijie Yan, Bo~Dai, and Dahua Lin.
\newblock Scene-aware generative network for human motion synthesis.
\newblock In {\em Proceedings of the IEEE/CVF conference on computer vision and pattern recognition}, pages 12206--12215, 2021.

\bibitem{wang2024exploring}
Yiqi Wang, Wentao Chen, Xiaotian Han, Xudong Lin, Haiteng Zhao, Yongfei Liu, Bohan Zhai, Jianbo Yuan, Quanzeng You, and Hongxia Yang.
\newblock Exploring the reasoning abilities of multimodal large language models (mllms): A comprehensive survey on emerging trends in multimodal reasoning.
\newblock {\em arXiv preprint arXiv:2401.06805}, 2024.

\bibitem{wang2024motiongpt}
Yuan Wang, Di~Huang, Yaqi Zhang, Wanli Ouyang, Jile Jiao, Xuetao Feng, Yan Zhou, Pengfei Wan, Shixiang Tang, and Dan Xu.
\newblock Motiongpt-2: A general-purpose motion-language model for motion generation and understanding.
\newblock {\em arXiv preprint arXiv:2410.21747}, 2024.

\bibitem{wei2022emergent}
Jason Wei, Yi~Tay, Rishi Bommasani, Colin Raffel, Barret Zoph, Sebastian Borgeaud, Dani Yogatama, Maarten Bosma, Denny Zhou, Donald Metzler, et~al.
\newblock Emergent abilities of large language models.
\newblock {\em arXiv preprint arXiv:2206.07682}, 2022.

\bibitem{weng2021diffusion}
Lilian Weng.
\newblock What are diffusion models?
\newblock {\em lilianweng. github. io}, page~21, 2021.

\bibitem{wu2025motionagent}
Qi~Wu, Yubo Zhao, Yifan Wang, Xinhang Liu, Yu-Wing Tai, and Chi-Keung Tang.
\newblock Motion-agent: A conversational framework for human motion generation with {LLM}s.
\newblock In {\em The Thirteenth International Conference on Learning Representations}, 2025.

\bibitem{wu2022saga}
Yan Wu, Jiahao Wang, Yan Zhang, Siwei Zhang, Otmar Hilliges, Fisher Yu, and Siyu Tang.
\newblock Saga: Stochastic whole-body grasping with contact.
\newblock In {\em European Conference on Computer Vision}, pages 257--274. Springer, 2022.

\bibitem{xu2023actformer}
Liang Xu, Ziyang Song, Dongliang Wang, Jing Su, Zhicheng Fang, Chenjing Ding, Weihao Gan, Yichao Yan, Xin Jin, Xiaokang Yang, et~al.
\newblock Actformer: A gan-based transformer towards general action-conditioned 3d human motion generation.
\newblock In {\em Proceedings of the IEEE/CVF International Conference on Computer Vision}, pages 2228--2238, 2023.

\bibitem{yang2023qpgesture}
Sicheng Yang, Zhiyong Wu, Minglei Li, Zhensong Zhang, Lei Hao, Weihong Bao, and Haolin Zhuang.
\newblock Qpgesture: Quantization-based and phase-guided motion matching for natural speech-driven gesture generation.
\newblock In {\em Proceedings of the IEEE/CVF Conference on Computer Vision and Pattern Recognition}, pages 2321--2330, 2023.

\bibitem{yu2020structure}
Ping Yu, Yang Zhao, Chunyuan Li, Junsong Yuan, and Changyou Chen.
\newblock Structure-aware human-action generation.
\newblock In {\em Computer Vision--ECCV 2020: 16th European Conference, Glasgow, UK, August 23--28, 2020, Proceedings, Part XXX 16}, pages 18--34. Springer, 2020.

\bibitem{zhang2023speechgpt}
Dong Zhang, Shimin Li, Xin Zhang, Jun Zhan, Pengyu Wang, Yaqian Zhou, and Xipeng Qiu.
\newblock Speechgpt: Empowering large language models with intrinsic cross-modal conversational abilities.
\newblock {\em arXiv preprint arXiv:2305.11000}, 2023.

\bibitem{zhang2023generating}
Jianrong Zhang, Yangsong Zhang, Xiaodong Cun, Shaoli Huang, Yong Zhang, Hongwei Zhao, Hongtao Lu, and Xi~Shen.
\newblock T2m-gpt: Generating human motion from textual descriptions with discrete representations.
\newblock In {\em Proceedings of the IEEE/CVF Conference on Computer Vision and Pattern Recognition (CVPR)}, 2023.

\bibitem{zhang2022motiondiffuse}
Mingyuan Zhang, Zhongang Cai, Liang Pan, Fangzhou Hong, Xinying Guo, Lei Yang, and Ziwei Liu.
\newblock Motiondiffuse: Text-driven human motion generation with diffusion model.
\newblock {\em arXiv preprint arXiv:2208.15001}, 2022.

\bibitem{zhang2023remodiffuse}
Mingyuan Zhang, Xinying Guo, Liang Pan, Zhongang Cai, Fangzhou Hong, Huirong Li, Lei Yang, and Ziwei Liu.
\newblock Remodiffuse: Retrieval-augmented motion diffusion model.
\newblock In {\em Proceedings of the IEEE/CVF International Conference on Computer Vision}, pages 364--373, 2023.

\bibitem{zhang2023finemogen}
Mingyuan Zhang, Huirong Li, Zhongang Cai, Jiawei Ren, Lei Yang, and Ziwei Liu.
\newblock Finemogen: Fine-grained spatio-temporal motion generation and editing.
\newblock {\em NeurIPS}, 2023.

\bibitem{zhang2024tinyllama}
Peiyuan Zhang, Guangtao Zeng, Tianduo Wang, and Wei Lu.
\newblock Tinyllama: An open-source small language model, 2024.

\bibitem{zhao2023large}
Zirui Zhao, Wee~Sun Lee, and David Hsu.
\newblock Large language models as commonsense knowledge for large-scale task planning.
\newblock {\em Advances in Neural Information Processing Systems}, 36:31967--31987, 2023.

\bibitem{zheng2024llamafactory}
Yaowei Zheng, Richong Zhang, Junhao Zhang, Yanhan Ye, Zheyan Luo, Zhangchi Feng, and Yongqiang Ma.
\newblock Llamafactory: Unified efficient fine-tuning of 100+ language models.
\newblock {\em arXiv preprint arXiv:2403.13372}, 2024.

\bibitem{zhu2023human}
Wentao Zhu, Xiaoxuan Ma, Dongwoo Ro, Hai Ci, Jinlu Zhang, Jiaxin Shi, Feng Gao, Qi~Tian, and Yizhou Wang.
\newblock Human motion generation: A survey.
\newblock {\em IEEE Transactions on Pattern Analysis and Machine Intelligence}, 2023.

\end{thebibliography}
